\documentclass[preprint,12pt,authoryear]{elsarticle}

\usepackage{amssymb}

\usepackage{soul,color}
\usepackage{multirow}
\usepackage{arydshln}   
\usepackage{hyperref}

\newcommand{\name}{\texttt{{CheckMate}}}
\newcommand{\dname}{\texttt{{CheckIt}}}

\usepackage{amssymb}
\usepackage{pifont}
\newcommand{\cmark}{\ding{51}}%
\newcommand{\xmark}{\ding{55}}%
\usepackage{subfig}
\usepackage{graphicx}

\renewcommand{\arraystretch}{1.3}

\journal{Information Processing and Management}

\begin{document}
\begin{frontmatter}




\title{Leveraging Social Discourse to Measure {\em Check-worthiness} of Claims for Fact-checking}

\author{Megha Sundriyal$^1$, Md Shad Akhtar$^1$, and Tanmoy Chakraborty$^2$}
\address{$^1$IIIT-Delhi, India, $^2$IIT-Delhi, India}
\ead{\{meghas, shad.akhtar\}@iiitd.ac.in, tanchak@iitd.ac.in}

\begin{abstract}
The expansion of online social media platforms has led to a surge in online content consumption. However, this has also paved the way for disseminating false claims and misinformation. As a result, there is an escalating demand for a substantial workforce to sift through and validate such unverified claims.
Currently, these claims are manually verified by fact-checkers. Still, the volume of online content often outweighs their potency, making it difficult for them to validate every single claim in a timely manner. Thus, it is critical to determine which assertions are worth fact-checking and prioritize claims that require immediate attention. 
Multiple factors contribute to determining whether a claim necessitates fact-checking, encompassing factors such as its factual correctness, potential impact on the public, the probability of inciting hatred, and more. Despite several efforts to address claim check-worthiness, a systematic approach to identify these factors remains an open challenge. To this end, we introduce a new task of fine-grained claim check-worthiness, which underpins all of these factors and provides probable human grounds for identifying a claim as check-worthy. We present \dname, a manually annotated large Twitter dataset for fine-grained claim check-worthiness. We benchmark our dataset against a unified approach, \name, that jointly determines whether a claim is check-worthy and the factors that led to that conclusion. We compare our suggested system with several baseline systems. Finally, we report a thorough analysis of results and human assessment, validating the efficacy of integrating check-worthiness factors in detecting claims worth fact-checking.
\end{abstract}


\end{frontmatter}


\section{Introduction}
\label{sec:intro}
The rapid evolution of online social media, driven by human communication needs and technological advancements, has transformed it from a mere leisure activity into a thriving business sector. Initially limited to desktop service, social media swiftly shifted to smartphones, becoming integral to our daily lives. As a result, it is now nearly impossible to carry out a day’s job without engaging with any form of social media. A recent report published in July’$22$ revealed that there were around $4.7$ billion social media users across the globe, accounting for $59$\% of the world’s total population \citep{kemp_2022}. Notably, this number has increased by $227$ million new users in the past year alone; this equates to a $5.1$\% annual growth rate -- equivalent to more than seven new users joining social media every second \citep{kemp_2022}. The freedom of speech and expression provided by these platforms has resulted in a tremendous increase in social media users. People are now empowered to openly convey their opinions and views on various topics without regard for political, religious, or economic restrictions. However, this unrestricted freedom comes at a high cost of factuality and accountability, allowing malicious users to propagate fake news, misinformation, and rumors without facing the consequences \citep{lazer2018science, zhang2020overview}.  

As a result, there has been a notable increase in the number of fact-checking organisations. On the other hand, the rate at which misinformation is generated and disseminated far outpaces the capacity of these manual fact-checkers. However, not all claims made online necessitate fact-checking attention. Unfortunately, fact-checkers often spend their valuable time reviewing such \textit{non-check-worthy} claims.  For example, \textit{``I'm having $102^o$C fever since last night. I'm immune to corona for sure."} is a claim; however, its veracity has no societal impact and thus does not require fact-checking. Table \ref{tab:checkworthy-examples} lists a few more examples. 

\begin{table*}[t]
    \centering
    \caption{A few examples of claims and their check-worthiness labels from different social media platforms.}
     \resizebox{\columnwidth}{!}
    {
    \begin{tabular}{p{32em}|c|c}
    \hline
    \hline
         \multicolumn{1}{c|}{\bf Text} & \bf Claim & \bf Check-worthy \\         
         \hline \hline 
         
         If you've swam in Burr Oak Lake, you're immune to the Coronavirus & No & - \\ \hline
         
        It's possible that a lab in the Chinese city of Wuhan was the source of the deadly animal-borne coronavirus that is now spreading throughout the world.  & Yes & Yes \\ \hline
        
         If this Corona scare doesn't end soon, I'm going to have to step in. & No & - \\\hline
         

         I'm having $102^o$C fever since last night. I'm immune to corona for sure. & Yes & No\\ \hline
         
         China already has a vaccine. They had it from the beginning. Now, they have the world at their mercy. & Yes & Yes \\ 
         \hline
         \hline
         
         
    \end{tabular}}
    \label{tab:checkworthy-examples}
\end{table*}

While false news might emanate accidentally, in most cases, it is actively spread by those who seek to benefit from it. One such instance is of $45th$ Presidential election in the United States. During this, the term \textit{`fake news'} on social media became highly prevalent \citep{grave2018learning}. The entire world witnessed the developing influence of fake news. It was reported that about $25$\% of Americans visited a fake news website, significantly impacting the election's outcome \citep{allcott2017social}. A similar problem was faced during the $2009$ Massachusetts special election for the United States Senate, where links to a website denigrating one of the candidates were effectively circulated by a network of fake Twitter accounts \citep{ratkiewicz2011detecting}. In response to such political events, the global number of fact-checking organizations expanded vastly. Various independent fact-checking organizations emerged in more than $50$ countries spanning nearly every continent that aim to verify claims \citep{cherubini_graves}. Even though fact-checking is typically done manually, particularly in journalism, the daily publication of millions of blog articles and social media posts makes manual fact-checking nearly absurd \citep{hoang2018location}. The misinformation and fake claims spread so quickly on online social media that they often outpace manual hoax-debunkers' strength, resulting in the proliferation of numerous unchecked erroneous and misleading claims over the internet. In the 1950s, \cite{toulmin2003uses} began the early work on argumentation mining. In his \textit{Argument Theory}, he defined the \textit{`claim'} as a statement that needs to be proven. He also stressed the term \textit{`claim'} as \textit{`an assertion that deserves our attention'}; while not very exact, it nonetheless provides a preliminary insight into claims' tasks. The claim notion is deeply embedded in the foundations of an argumentation task.

\paragraph{\textbf{Motivation}} Since COVID-19 hit the world in late $2019$, misinformation and false claims have been on the loose again. 
Internet hits increased by $50$\% to $70$\% in $2020$, with streaming rising by at least $12$\% \citep{beech_2020}. Unfortunately, this boosted engagement on online social media platforms furnished opportunities for vicious users to spread misleading and false information. Around the same time, numerous social media posts spreading misinformation about the virus began to appear, ranging from remedies to symptoms to anti-vaccine campaigns. As the coronavirus spread, so did the conjecture about its cure. For example, misinformation about the ability of alcohol/disinfectants to kill coronavirus resulted in a considerable increase in methanol-induced mortality. The Ministry of Health of Iran reported $5011$ people were poisoned with methanol poisoning as of April $27$, $2020$, with $505$ verified deaths \citep{sefidbakht2020methanol}. The overall number of reported fatalities from February to April $2020$ was nearly eight times that of the same period in $2019$; however, reports before $2019$ reveal fewer deaths due to methanol intoxication \citep{sefidbakht2020methanol}. Such baseless claims had unprecedented repercussions, causing monetary and irreplaceable human life loss, thus needing to be fact-checked for their truthfulness. Unfortunately, the volume of online claims often outweighs the potency of manual fact-checkers, making it difficult for them to validate every claim produced online. Thus, identifying check-worthy claims is the foremost step of any fact-checking pipeline. It is critical to determine which assertions are worth fact-checking and to verify such claims that require immediate attention first. 

\begin{figure}[t]
    \centering
    \includegraphics[scale=0.72]{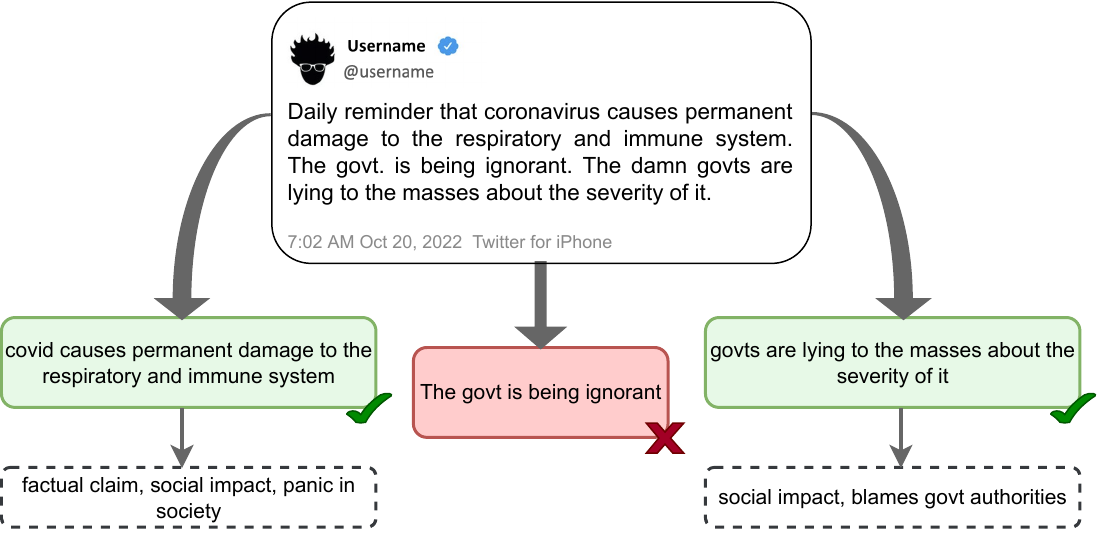}
    \caption{A motivating example for the task of fine-grained check-worthy claim detection. The green line (tick) denotes check-worthy claims, and the red line (cross) represents non-check-worthy claims. The white block (dashed) contains justifications for check-worthy claims.}
    \label{fig:example}
\end{figure}

Clinching the veracity of a claim manually is not only time-consuming but also intellectually demanding. Thus, natural language processing and machine learning communities have recently begun to confront this formidable issue automatically \citep{vlachos-riedel-2014-fact, thorne-etal-2018-fact, atanasova2019overview, das2023state}. While better models for claim check-worthiness detection are continuously being developed, there is little research on the explainability aspects of these decisions. There hasn't been any systematic effort yet to pinpoint the contributing components and understand how they relate to the check-worthiness of a claim. Before defining a claim as worth checking, several elements must be considered, such as determining how factual the claim is, whether the assertion impacts the general public, and so on. Consider the example in Figure \ref{fig:example}, for the task of binary claim check-worthiness identification at the coarse level, the entire tweet will be marked as a check-worthy. On closer study, however, we discover three claims, each with a distinct significance. The text pieces  \textit{``covid causes permanent damage to the respiratory system and immune system"} and \textit{``govts are lying to the masses about the severity of it"} represent the statements in the tweet that are worth checking, with plausible justifications in dashed boxes. \textit{``The govt is being ignorant''} on the other hand, is a non-check-worthy claim that does not necessitate fact-checking. The task of fine-grained claim check-worthiness identification has numerous advantages. It gives a preference scale, allowing fact-checkers to prioritize critical fact-checking assertions without digging through millions of claims. Additionally, it adds the explainability perspective to binary claim check-worthiness identification.

\paragraph{\textbf{Our Contributions}} We make the following contributions through this work:
\begin{itemize}
\item To mitigate the non-availability of a labeled corpus for fine-grained claim check-worthiness detection in Twitter. We create a large, manually annotated Twitter dataset, \dname, with over $5.9k$ claims, integrating the viewpoints of the general public, fact-checkers, and the government.

\item We propose \name, a unified approach that jointly considers a claim's check-worthiness and the rationale that contributed to it. It amalgamates linguistic features through part-of-speech and dependency tags and contextual attributes through a fine-tuned language model.

\item We evaluate \name\ against multiple state-of-the-art claim check-worthiness models. We show significant performance improvements over baseline systems across various evaluation metrics. 
\end{itemize}

The remaining paper is structured in the following manner. Section \ref{sec:related-work} goes over related work. The dataset description is given in Section \ref{sec:dataset}. In Section \ref{sec:methodology}, the task is formally defined along with the detailed proposed methodology. Section \ref{sec:experimental-settings} overviews the baseline systems and discusses the assessment process and experimental settings. Section \ref{sec:results-and-analysis} offers the results and analysis, followed by the conclusion in Section \ref{sec:conclusion}.

\section{Related Work}
\label{sec:related-work}
Through rapid information release and equal consumption, social media has greatly aided the democratization of information distribution. Simultaneously, because of the lag in information supervision, fake news (including sarcasm, misinformation, rumors, and so on) has become rampant and flooded, severely affecting people's lives, social stability, and even national security. Misinformation is now widely disseminated on social media through false and misleading claims, which has increased interest in combating false or deceitful information in academia and industry. Over $200$ fact-checking organizations worldwide, including PolitiFact, FactCheck, Snopes, and Alt News,\footnote{\url{https://www.politifact.com/}, \url{https://www.factcheck.org/}, \url{https://www.snopes.com/}, \url{https://www.altnews.in/}} have launched initiatives to perform manual fact-verification. Unfortunately, these efforts are inadequate, given the magnitude of misinformation disseminated through multiple online communication channels. To combat this deluge of misinformation, the natural language processing and machine learning communities have begun to address the issue of fact-checking through automated means \citep{barron2018overview, atanasova2019overview, barroncedeno2020overview, shaar2021overview}. Recent studies have focused on various subtasks around fact-checking ranging from automatic identification of claims \citep{levy2014context, levy2017unsupervised, gupta2021lesa, sundriyal2021desyr}, to extracting evidence to establish claim veracity \citep{thorne2018fever, barron2018overview, barron2020overview, shaar2021overview, sundriyal-etal-2022-document}. It isn't easy, not only for humans but also for automated systems, to fact-check every claim, given the daily magnitude of claims proffered online. This necessitates systems segregating check-worthy content online, saving the valuable time and resources of manual fact-checkers and automated systems. The first and most significant step in fact-checking is to determine whether a piece of text makes \textit{``an assertion about the world could be checked or not"} \citep{konstantinovskiy2021toward}. The task of identifying check-worthy claims has drawn the attention of numerous researchers \citep{hassan2015detecting, gencheva2017context, jaradat2018claimrank, vasileva2019takes}. Until now, studies on claim check-worthiness have primarily been divided into the following two categories.

\paragraph{\textbf{Check-worthy claims in political discourse and debates}} The first category focuses on identifying check-worthy claims in debates or political speeches. With a pioneering attempt by \cite{hassan2017claimbuster}, claim check-worthiness detection has emerged as a major research area in recent years. They developed, ClaimBuster, the first-of-its-kind system to target a claim's check-worthiness in a debate. It was trained on a massive manually annotated dataset of US election debates from $1960$ to $2012$, totaling $30$ debates and $28,029$ transcribed sentences. Each statement made during a political discussion was categorized into one of three categories: non-factual, unimportant factual, or check-worthy factual. Focusing on the $2016$ US Presidential debates,  Gencheva \textit{et al.} \cite{gencheva2017context} obtained binary annotations for check-worthiness from various fact-checking organizations. Later they developed ClaimRank, which was trained on additional data, including Arabic content. To determine whether a sentence would be selected for fact-checking, Patwari \textit{et al.} \cite{patwari2017tathya} used a model that is similar to boosting. Vasileva \textit{et al.} \cite{vasileva2019takes} used a multi-task learning neural network. The task was to predict whether a sentence would be selected for fact-checking by each of nine different fact-checking organisations. The CheckThat! Lab also organized similar shared tasks in the year $2018$ \cite{barron2018overview} and $2019$ \cite{atanasova2019overview}, focusing on political debates and speeches. The attention of several researchers was drawn to these shared tasks. In the $2018$ edition, seven teams submitted their runs for Task 1 on check-worthiness using systems based on word embeddings and RNNs \cite{ghanem2018upv, agez2018irit, hansen2018copenhagen}. Eleven teams submitted strategies for the corresponding task in the $2019$ edition. Most of them utilized RNNs and experimented with various intriguing representations \cite{ hansen2019neural, altun2019tobb, coca2019checkthat, dhar2019hybrid, favano2019theearthisflat}. 

Research on estimating claim-worthiness in debates and political speeches has been thoroughly studied and made substantial progress. Due to the lack of structure and proper linguistic properties, these approaches do not work well for noisy texts. The social media expansion has boosted preponderance of false and misleading claims; as a result, current research has turned to identify claim check-worthiness in social media.

\paragraph{\textbf{Check-worthy claims in social media posts}} In recent years, claim check-worthiness detection has escalated great demand. To address this, in their recent editions, CheckThat! Labs organized shared tasks on identifying check-worthy claims with a primary focus on social media texts \cite{barroncedeno2020overview,  nakov2021overview, nakov2022overview}. These shared tasks attracted multiple systems modeled to handle noisy text from social media platforms. The $2020$ edition featured three central tasks: detecting previously fact-checked claims, evidence retrieval, and actual fact-checking of claims. Along with English, they offered the tasks in Arabic and Spanish. For Arabic, several teams improved pre-trained models, such as AraBERT \cite{antoun2020arabert} and multilingual BERT \cite{williams2020accenture, hasanain2020bigir}. In case of the English task, multiple systems harnessed the strength of pre-trained Transformers, specifically BERT and RoBERTa \cite{williams2020accenture, hasanain2020bigir, cusmuliuc2020uaics, kartal2020tobb}. Other methods extracted tweet embeddings from tweets using pre-trained models like GloVe and Word2vec, which were then fed into a neural network or an SVM \cite{cheema2020check_square}. The top-ranked system in the $2021$ edition also leveraged transformer-based models \cite{nakov2021overview, williams2021accenture, zhou2021fight}. The first shared task in the $2022$ edition was to anticipate which Twitter posts should be fact-checked, with an emphasis on COVID-19 and politics. It was put forth in Arabic, Bulgarian, Dutch, English, Spanish, and Turkish.  A total of $19$ teams competed, with the majority of submissions achieving significant improvements over baselines using Transformer-based models such as BERT and GPT-3 \cite{nakov2022overview, toraman2022arc, agrestia2022polimi}. Alam \textit{et al.} \cite{alam2021fighting} created a multi-question annotation schema of COVID-19 tweets structured around seven claim check-worthiness questions.

 Subjectivity of annotations for existing data on claim check-worthiness detection is a well-known issue \cite{konstantinovskiy2021toward}. 
While better models for detecting claim check-worthiness are constantly being developed, there is a lack of literature on explainability aspects of these binary decisions. Inspired by the annotation schema proposed by Alam \textit{et al.} \cite{alam2021fighting}, we take it a step further and attempt to back the binary label decision with six auxiliary labels, forging a new task of fine-grained claim check-worthiness.

\section{Dataset}
\label{sec:dataset}
Despite the abundance of claims in online social media, the literature does not indicate any significant attempt in fine-grained check-worthy claim identification, identifying the rationales behind marking a claim check-worthy. To fill this void, we propose \dname, a large-scale manually annotated Twitter corpus with fine-grained claim check-worthiness labels. This section describes our proposed dataset and annotation process in detail. 

\subsection{Dataset Selection and Annotation}
Several check-worthy claim detection datasets have been released in recent years \citep{jaradat2018claimrank, gencheva2017context, nakov2021overview, barroncedeno2020overview,  shaar2021overview}; nevertheless, each of them classifies the entire sentence or document as check-worthy and do not furnish fine-grained check-worthiness labels that provide reasonable grounds to quantify a claim as check-worthy. We discover that CURT, a recently released Twitter dataset by \cite{sundriyal2022empowering}, provides claim units in a specific tweet and is the best fit for our task. According to our observations, each claim unit within a tweet has a different relevance and must be validated separately. This prompted us to annotate each claim unit rather than the entire tweet for our fine-grained claim check-worthiness task. As a result, we use the CURT dataset \citep{sundriyal2022empowering} and annotate it for fine-grained claim check-worthy labels. The annotation task consists of two steps: (a) creating fine-grained labels that consider several factors, including whether a tweet contains a factual assertion, whether it is harmful to society, etc., and (b) determining whether a claim is check-worthy or not based on the previously mentioned factors. Inspired by \cite{alam2021fighting}, we choose the following binary questions to reflect the grounds for check-worthiness and formulate them into six binary fine-grained check-worthiness labels.

\begin{enumerate} 
    \item[\textbf{\textit{L1.}}] \textbf{Is there a factual claim in the tweet that can be verified?} The questions asks for the tweets that provide definitions, mention some statistics in the present or past, make reference to laws, etc. 
    
    \item[\textbf{\textit{L2.}}] \textbf{Does it seem like the tweet contains false information?} This question solicits a personal opinion. It does not inquire whether the assertion in the tweet is factually accurate; rather, it asks whether the claim \textit{seems} to be incorrect.
    
    \item[\textbf{\textit{L3.}}] \textbf{Will the tweet have an impact on general public or pique the public's interest?} In general, claims containing information about potential COVID-19 cures, updates on the number of cases, government measures, or discussing rumours and spreading conspiracy theories are of general public interest. 
    
    \item[\textbf{\textit{L4.}}] \textbf{Is the tweet harmful to the society?} This question also asks for an objective judgment: to identify tweets that can cause potential harm, as such tweets need to be verified quickly and hence making them check-worthy. 
    
    \item[\textbf{\textit{L5.}}] \textbf{Should the claim in the tweet be verified by a professional fact-checker?} This question solicits a personal opinion. It emphasizes the need for a qualified fact-checker to confirm the assertion, ruling out straightforward claims for a novice to verify. 
    
    \item[\textbf{\textit{L6.}}] \textbf{Should the tweet get the attention of a government entity?} This question solicits the judgement on whether the target tweet should be brought to the attention of a government entity or policymakers in general.
\end{enumerate}

We progressed through multiple iterations of refinements to attune better to the dataset. At each iteration, $100$ random tweets were annotated by three annotators. The annotators settled the ambiguous cases collectively, while the unresolved claims that demanded guideline revisions were corrected in later iterations. We examined all previous annotations for each modification in the guideline to ensure that the annotations mirrored the most advanced version of the annotation guidelines. Utilizing the final guidelines, an additional batch of $100$ tweets were annotated as part of the last sprint of the pilot annotation. We obtain an average inter-annotator agreement score of $0.73$ using the Cohen Kappa \cite{cohen1960}. Some examples of annotated claims are shown in Table \ref{tab:annotated-samples}. The first example contains two check-worthy claims, but the contributing factors for their worthiness vary considerably. A closer examination reveals that the latter claim has a more detrimental effect on society than the former. The second tweet contains three claims, only two of which are worth checking. The first one, \textit{``\#ozone is being used to destroy \#COVID-19,"} is a verifiable factual claim that is of great public interest. It should be fact-checked as quickly as possible as it discusses COVID-19 protection. The second claim also includes a verifiable factual claim, but it may not be of broad interest to the general public. Similarly, the third tweet contains two check-worthy claims and one non-check-worthy claim.

\begin{table*}[th]
    \centering
    \renewcommand*{\arraystretch}{1.0}
    \caption{A few sample tweets with claim spans from our dataset labeled for claim check-worthiness and fine-grained claim check-worthiness labels. CW denotes check-worthy label.}
    
    \resizebox{\textwidth}{!}{
    \begin{tabular}{p{15em}|p{12em}|c|p{12em}}
    \hline
    \hline
    \textbf{Tweet} & \textbf{Claim} & \textbf{CW} & \textbf{Fine-grained labels}\\ \hline \hline
    
    \multirow{2}{15em}{Healthy people don't spread \#covid\_19, people who have recovered and are immune don't spread the virus. You know who else doesn't spread the virus? People wearing masks. \#Masks4All} & Healthy people don't spread \#covid\_19 &  \cmark & \textit{{false information, interests general public}} \\ \cdashline{2-4} 
    & people who have recovered and are immune don't spread the virus & \cmark & \textit{verifiable factual claim, false information, impacts the general public, harmful for society, fact-checker should verify}\\ 
    \hline

    \multirow{3}{15em}{\# ozone is being used to destroy \#COVID-19 ps. Vaccines for RNA viruses cause mutation, so if you want a worse pandemic than this or to die from an injection, that’s all that will be on offer. Right now, there’s a known cure.} &  \#ozone is being used to destroy \#COVID-19  &  \cmark & \textit{verifiable factual claim,	false information, impacts general public, harmful for society, fact-checker should verify}\\ \cdashline{2-4} 
    & Vaccines for RNA viruses cause mutation & \cmark  & \textit{verifiable factual claim, fact-checker should verify} \\ \cdashline{2-4}
    & there is a known cure & \textcolor{red}{\xmark} & 
    \\ \hline
    
    \multirow{3}{15em}{@username From what I’ve read, the coronavirus is no more dangerous than the common cold or the seasonal flu. Symptoms are usually so mild that one doesn’t even notice they are sick. The people who run into problems are the immune-compromised (like the very old or the already ill)} & coronavirus is no more dangerous than the common cold or the seasonal flu & \cmark & \textit{verifiable factual claim, false information, impacts the general public, harmful for society, fact-checker should verify}
    \\ \cdashline{2-4}
    & Symptoms are usually so mild that one does not even notice they are sick & \cmark & \textit{verifiable factual claim, impacts the general public, fact-checker should verify}
    \\ \cdashline{2-4}
    & The people who run into problems are the immune-compromised & \textcolor{red}{\xmark}
    \\
    \hline
    \hline
    \end{tabular}}
    \label{tab:annotated-samples}
\end{table*}

\subsection{Dataset Statistics}
For the fine-grained claim check-worthiness identification task, we annotated $5920$ claims in total. We split \dname\ into three sections: training set, validation set, and test set, with a $70$:$15$:$15$ split. The summary of dataset statistics is shown in Table \ref{tab:dataset_stats}. Table \ref{tab:dataset_stats_aux} depicts the statistics on six auxiliary labels used to annotate the samples for fine-grained check-worthiness labels. Every auxiliary question has a broadly consistent distribution, except for $L4$ and $L6$, where the distribution is slightly skewed towards \textit{No}. We can see that more claims fall under the category of not harmful compared to harmful in $L4$, which asks whether the claim is harmful to society. Similarly, most claims in $L6$ do not necessitate the attention of government entities.

\begin{table}[t]
\centering

\caption{Dataset statistics of our dataset, \dname.}
\resizebox{0.5\textwidth}{!}
{
\begin{tabular}{lcc}
\hline
\hline
\textbf{Dataset} & \textbf{Check-worthy} & \textbf{Non check-worthy} \\  \hline \hline
Training & 2677  & 1467 \\
Validation  & 574   & 314 \\
Testing & 573   & 315 \\ 
Total & 3824 & 2096 \\ \hline \hline
\end{tabular}}
\label{tab:dataset_stats}
\end{table}
\begin{table}[t]
\centering
\caption{Statistics on six auxiliary fine-grained labels ($L1$–$L6$) used to annotate the samples for fine-grained check-worthiness labels. The columns refer to the total number of times the question was answered, yes or no, in the corresponding dataset. Also, note that one instance can apply to many questions because these questions are not mutually exclusive.}
\resizebox{0.45\textwidth}{!}{
\begin{tabular}{ccccccc}
\hline
\hline \multirow{2}{*}{\textbf{Label}} & \multicolumn{2}{c}{\textbf{Training}}  & \multicolumn{2}{c}{\textbf{Validation}}    &
\multicolumn{2}{c}{\textbf{Testing}} 
\\
\cline{2-7}

& \multicolumn{1}{c}{\textbf{Yes}} & \multicolumn{1}{c}{\textbf{No}} & \multicolumn{1}{c}{\textbf{Yes}} & \multicolumn{1}{c}{\textbf{No}} & \multicolumn{1}{c}{\textbf{Yes}} &
\multicolumn{1}{c}{\textbf{No}} \\ 
\hline
\hline

L1    & 2336    & 1808 &                           509        & 379                      & 516 & 372                              \\
L2    & 2535     & 1609                           & 602          & 286                    & 662   & 226                            \\
L3    & 2910   & 1234                             & 609                & 279                & 630 & 258                               \\
L4    & 1690   & 2454                            & 358            & 530                  & 438   & 450                            \\
L5   & 2076   & 2068                            & 380                 & 508             & 443    & 445                           \\ 
L6  & 1711 & 2433 & 368 &  520  & 416 & 472 \\ \hline \hline
\end{tabular}}

\label{tab:dataset_stats_aux}
\end{table}

\subsection{Comparison with existing claim datasets}
Existing datasets in computational argumentation cover a wide range of aspects of claim analysis. Table \ref{tab:comparison_datasets} compare our proposed dataset \dname\ with other publicly available claim-related datasets. There are a variety of datasets available for identifying claims in a document, verifying claims, indicating claims that need to be fact-checked, and so on \citep{peldszus-stede-2015-joint, stab-gurevych-2017-parsing, thorne-etal-2018-fact, gupta2021lesa, sundriyal2022empowering}. The literature, however, lacks corpora with fine-grained claim check-worthiness at the claim span level. CheckThat! Labs organized a shared task in CLEF, in which one prominent resource in the realm of claim check-worthiness was provided. The dataset provided here is smaller and does not include justifications for the check-worthiness label. However, our proposed dataset, \dname, provides fine-grained check-worthiness labels to approximately $5.9k$ claims, making it the first dataset of its kind. 

\begin{table*}[ht]
    \caption{Description of existing benchmark claim datasets and task compared with our proposed dataset, \dname. \textbf{\#} denotes the total number of instances, while ADU denotes Argumentative Discourse Unit.}
    \centering
    \resizebox{\textwidth}{!}{
    \begin{tabular}{lcccc}
    \hline
    \hline
    \textbf{Dataset} & \textbf{\#} & \textbf{Source}  & \textbf{Granularity} & \textbf{Task} \\ 
    \hline
    \hline

    German Microtext \citep{peldszus-stede-2015-joint} & 112 & Short texts & Sentence & ADU Identification\\
    Student Essays \citep{stab-gurevych-2017-parsing} & 90 & Student essays & Document & Claim Detection \\ 
    FEVER \citep{thorne2018fever} & 185k & Wikipedia & Sentence & Claim Verification \\
    MultiFC \citep{augenstein2019multifc} & 36534 & Fact-checking sites & Sentence & Claim Verification \\
    LESA \citep{gupta2021lesa} & 9894 & Tweets & Sentence & Claim Detection \\ 
    CURT \citep{sundriyal2022empowering} & 7555 & Tweets & Phrase & Claim Span Identification \\
    CLEF-2019 Task 1 \citep{barron2018overview} & 23k & Political debates &  Sentence & Claim Check-worthiness \\ 
    CLEF-2020 Task 1 \citep{barroncedeno2020overview} & 962 & Tweets &  Sentence & Claim Check-worthiness \\ 
    CLEF-2021 Task 1A \citep{shaar2021overview} & 1312 &  Tweets &  Sentence & Claim Check-worthiness \\
    CLEF-2022 Task 1A \citep{nakov2022overview} & 3040 &  Tweets & Sentence & Claim Check-worthiness \\
    \hline
    \dname\ & 5920 & Tweets & Phrase & Fine-grained Check-worthiness\\
    \hline
    \hline
    \end{tabular}}
    \label{tab:comparison_datasets}
\end{table*}

\section{\name: Our Proposed System}
\label{sec:methodology}
\begin{figure*}[t]
    \centering
    \includegraphics[width=\textwidth]{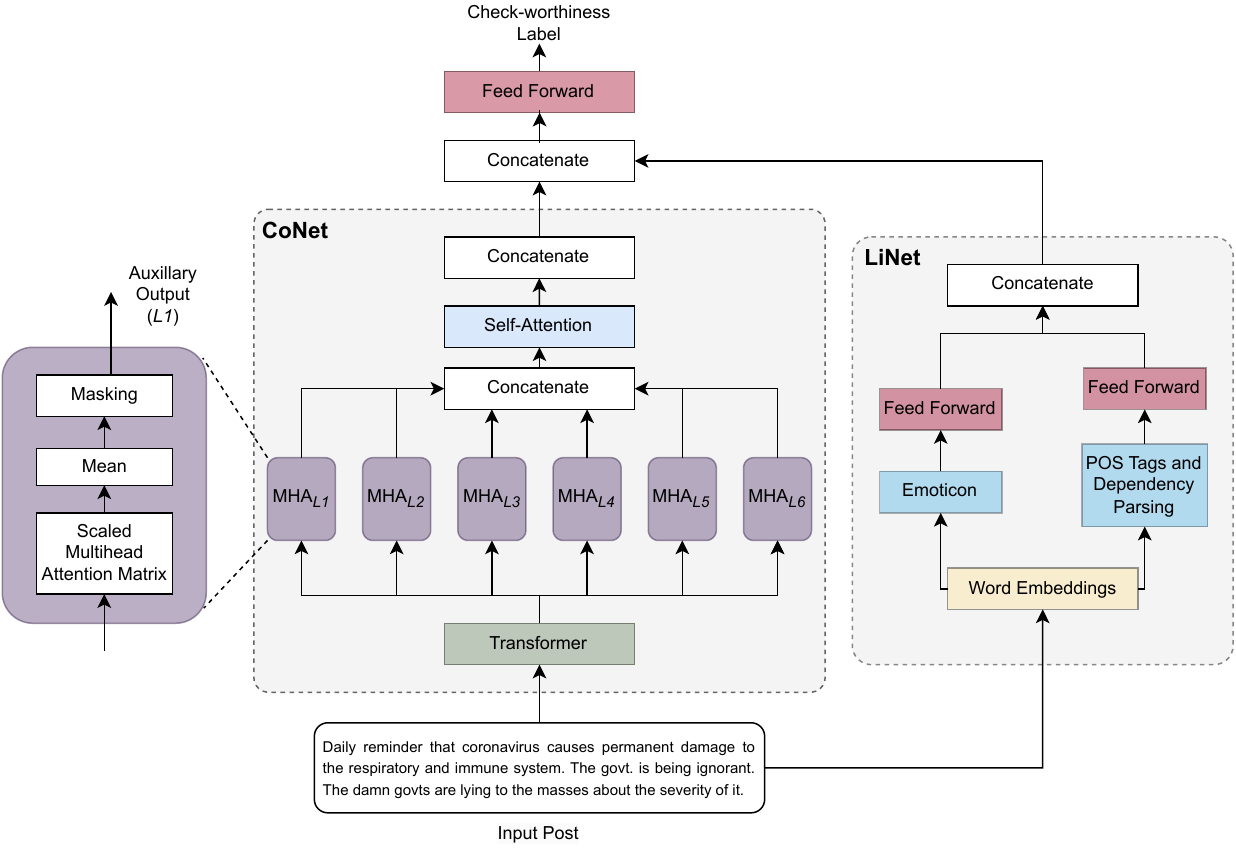} 
    \caption{The architecture of our proposed model, \name, for detecting fine-grained claim check-worthiness. The model is divided into two modules: CoNet and LiNet. CoNet aims to capture contextual features of the input text, whereas LiNet seeks linguistic features. MHA denotes Multihead Attention Block corresponding to each Auxiliary Label ($L1$-$L6$).
    }
    \label{fig:model_architecture}
\end{figure*}

We witness that before $2020$, the narrative on claim check-worthiness detection is primarily contextual or linguistic. To achieve our goal of utilizing both, we propose an integrated model, \name. It seeks to combine linguistic features obtained from part-of-speech (POS) tags and dependency trees with contextual factors extracted from the transformer-based model, BERT \cite{devlin2018bert}.  
A high-level architecture of our proposed model is shown in Figure \ref{fig:model_architecture}. It has two main components -- CoNet consisting of the BERT framework, a module to optimize six attention heads of BERT (one attention head per question) and obtain the contextual features, and LiNet to incorporate linguistic features emanating from POS tags and dependency trees. The following subsections provide a formal task definition of the fine-grained extension of the claim check-worthiness problem and intricate details of the aforementioned modules of the proposed model. 

\subsection{Task Definition}
The primary task is binary text classification, determining whether a piece of text is worth fact-checking. We strive to enhance the claim check-worthiness objectives with additional questions that would aid in defining the rationale for the binary check-worthiness label. Inspired by  Alam \textit{et al.} \cite{alam2021fighting}, we include six auxiliary labels that align most with the final claim check-worthy label, keeping in sight the viewpoints of journalists, policymakers, society, and manual fact-checkers. 

Formally, let $D = \{(x_{1}, y_{1}, A_{1}), (x_{2}, y_{2}, A_{2}),..., (x_{m}, y_{m}, A_{m})\}$ be the set of $m$ labelled instances, where $x_i$ denotes the input claim, $y_i \in \{0, 1\}$ denotes the primary claim check-worthy label where $1$ denotes that the claim is worth fact-checking and $0$ denotes not check-worthy. $A_{i} = \{a_{i1}, a_{i2}, a_{i3}, a_{i4}, a_{i5}, a_{i6}\}$ represents the set of auxiliary labels. The value of each $a_i \in \{0, 1\}$ for $i \in \{1, 6\}$. Let $x$ be a post in the dataset such that $x = \{w_{1}, w_{2},...,w_{n}\}$ comprising of $n$ words. The objective is to build a multi-label classifier $F$ that maps an input text instance $x$ to its associated primary label $y$ and set of auxiliary labels $A$, while optimizing some evaluation metrics.

\subsection{Model Architecture}
Figure \ref{fig:model_architecture} architecture depicts the high-level architecture of our proposed unified model, \name, for fine-grained claim check-worthiness. The model consists of two primary modules: a Contextual Network (CoNet), a transformer-based architecture, BERT, intended to extract contextual elements of the text while also optimizing the six attention heads of BERT corresponding to the six auxiliary fine-grained questions, and a Linguistic Network (LiNet), which takes into account linguistic features retrieved from part-of-speech (POS) tags and dependency trees (Dep).

\subsubsection{Contextual Network (CoNet)}
To extract the contextual features of the input claim, the CoNet module includes a BERT layer followed by a self-attention layer. We begin by pre-training BERT with our claim check-worthy corpus and then compute six attention vectors using BERT's six attention heads to fine-tune BERT during training. We hypothesize that assigning one attention head to each auxiliary label will assist the model in catering to each label specifically. 

To begin with, we tokenize every $i^{th}$ input text $x_i=\{w_{1}, w_{2},...,w_{m}\}$. We then utilize the pre-trained BERT model to obtain hidden representations for the $i^{th}$ post, $\hat{x_i} \in \mathbb{R}^{M \times d} $ where $M$ is the maximum sequence length and $d$ is the feature dimension. Let $A = \{A^{1}, A^{2},...,A^{6}\}$ be the set of query-key attention matrices of BERT for post $\hat{x}$, where $A^{l} \in A$ denotes the $l^{th}$ query-key attention matrix for post $x$, corresponding to the $l^{th}$ fine-grained label derived using one of the six auxiliary questions. We then compute the mean of the query attention scores, produce a final attention vector, and use a masked softmax function to normalize it. Finally, we optimize the label-wise supervised loss for each post between the fine-grained gold label and the corresponding predicted label.

\subsubsection{Linguistic Network (LiNet)}
The LiNet module seeks to extract linguistic information from the input text by employing two critical linguistic constructs: part-of-speech tags and dependency parse trees. For each $i^{th}$ input text $x_i = \{w_1, w_2,...,w_m\}$, we acquire a series of associated POS tags, yielding the set $\{p_1, p_2,...,p_m\}$. Because of the limited number of POS tags available in a language, the underlying concern of this modelling approach is the small constricted vocabulary. After that, dependency parsing is used to abstract the input text's grammatical structure. It discovers the related words and derives a directed relation, $d(w_j, w_k)$, between any two tokens $w_j$ and $w_k$, in which $w_j$ is the head and $w_k$ is the dependent.  These dependency relations are obtained using Python's SpaCy library.\footnote{www.spacy.io} Along with the current tokens, we also capture the dependency tags of their headwords as well. Furthermore, to benefit from their semantics, we encode emoticons present in the input post. 

Following this, we integrate contextual features retrieved from self-attended vectors of CoNet along with the linguistic features obtained from LiNet. The concatenated vector is fed into two multi-layered perceptrons for the final classification. The objective is to combine various characteristics from distinct viewpoints and leverage them for the final classification. The predictions of six auxiliary labels are then made using six sigmoid neurons with binary cross-entropy loss.

\section{Experimental Settings}
\label{sec:experimental-settings}
In this section, we outline the baseline systems and evaluation metrics that we employ to contrast the effectiveness of the proposed model and baselines. 

\subsection{Evaluation Measures}
For coarse-grained check-worthiness detection, we opt for macro and weighted F1-score as the primary evaluation metrics. To better comprehend the performance of the systems to detect check-worthy claims, we also report class-wise precision, recall, and F1-score. We hinge on weighted F1-scores corresponding to each fine-grained label for a fine-grained setup.

\subsection{Baseline Models}
Every year for the last three years, the CLEF shared tasks have been introduced to detect check-worthy claims in social media posts \citep{ atanasova2019overview, barroncedeno2020overview, barron2018overview, shaar2021overview}. These shared tasks have attracted numerous systems in the past years. To assess and contrast the performance of our proposed system, \name, we consider these systems as our baselines. We compare performance in two distinct setups. In the first setup, coarse-grained, we compare the results for binary claim check-worthiness labels against existing state-of-the-art claim check-worthiness systems models. For the second set-up, fine-grained, six auxiliary questions ($L1$-$L6$) are deemed independent tasks compared with several baseline systems. We use the following baseline systems to compare our proposed model's performance. 

\begin{itemize}
    \item \textbf{AI Rational} \citep{savchev2022ai}: The first rank at CLEF-2022 was bagged by the team AI Rational. They refined several pre-trained transformer models, including BERT \citep{devlin2018bert}, DistilBERT \citep{sanh2019distilbert}, and RoBERTa \citep{liu2019roberta}. Their best model, however, made use of RoBERTa-large. 
    
    \item \textbf{PoliMi FlatEarthers} \citep{agrestia2022polimi}: The authors used a pre-trained GPT3 \citep{brown2020language} model and fine-tuned it for the CLEF-2022 challenge; they ranked third for the shared task of claim check-worthiness.
    
    \item \textbf{NLPIR@UNED} \citep{martinez2021nlp}: The authors proposed a transformer-based approach that ranked first in task 1 at CLEF-2021. They reported BERTweet \citep{nguyen2020bertweet} as their best submission.
    
    \item \textbf{Fight for 4230} \citep{zhou2021fight}: The runner-up of the CLEF-2021 challenge utilized various preprocessing techniques, including normalizing punctuation to English, removing all links and non-ASCII, etc. Their best submission leveraged BERTweet architecture along with a dropout and a classifier layer. 
    
    \item \textbf{Accenture} \citep{williams2020accenture}: Team Accenture bagged first position in the shared task 1 in CLEF-2020. The best model of the three models they suggested (BERT \citep{devlin2018bert}, BERTweet \citep{nguyen2020bertweet}, and RoBERTa \citep{liu2019roberta}) was RoBERTa \citep{liu2019roberta} with additional pooling and classification layer.
    
    \item \textbf{Team Alex} \citep{nikolov2020team}: The system placed second in the CLEF-2020 challenge by using preprocessing techniques such as splitting hashtags into separate words, unifying COVID-19 hashtags, replacing `@' with `user,' and so on. Furthermore, two additional methods were tested: Corona Pre-Processing and SW+C Preprocessing.
    
    \item \textbf{BERT} \citep{devlin2018bert}: We fine-tune BERT, a bidirectional transformer-inspired auto-encoder language model, for our task of check-worthiness detection.  
    
    \item \textbf{RoBERTa} \citep{liu2019roberta}: RoBERTa is a robustly optimized variant of BERT with improved training methodology. Like BERT, we also fine-tune RoBERTa for our downstream task of claim check-worthiness. 
    
    \item \textbf{XLNet} \citep{yang2019xlnet}: XLNet is also a bidirectional transformer-inspired language model like BERT, the only difference being that the latter is an auto-regressive language model.
    
    \item \textbf{BERTweet} \citep{nguyen2020bertweet}: BERTweet is a large-scale language model pre-trained for English Tweets. It is trained based on the RoBERTa pre-training methodology and uses a huge corpus of $850$M English Tweets. 
    
\end{itemize}

Apart from the existing state-of-the-art systems, we also explore several traditional machine learning models, like Logistic Regression (LR), Multinomial Naive Bayes (MNB), Random Forest (RF), and Support Vector Machine (SVM). We leverage the pre-trained model of Word2vec \citep{mikolov2013efficient} to obtain embeddings for the tweets, which we feed into these models.


\begin{table*}[th]
\centering
\renewcommand*{\arraystretch}{1.2}
\caption{Performance comparison of \name\ on the binary, coarse-grained claim check-worthiness task against several baseline systems using our dataset, \dname. We report precision (P), recall (R), and F1-scores (F1) for both classes and the Accuracy. The \textbf{bold} type indicates the best performance, while \underline{underlined} numbers represent the second-best results for each metric. $\Delta$ represents the difference between the \name\ scores and the best-performing baseline across that evaluation measure.}

\resizebox{\textwidth}{!}{
\begin{tabular}{lccccccccc}
\hline
\hline
\multirow{2}{*}{\textbf{Model}} & \multicolumn{1}{c}{\multirow{2}{*}{\textbf{Accuracy}}} & \multicolumn{2}{c}{\textbf{F1-score}}  & \multicolumn{3}{c}{\textbf{Check-worthy}}  & \multicolumn{3}{c}{\textbf{Non-check-worthy}}                                                                 \\ \cline{3-10} & 
\multicolumn{1}{c}{} & \multicolumn{1}{c}{\textbf{Macro}} & \multicolumn{1}{c}{\textbf{Weighted}} & \multicolumn{1}{c}{\textbf{P}} & \multicolumn{1}{c}{\textbf{R}} & \multicolumn{1}{c}{\textbf{F1}} & \multicolumn{1}{c}{\textbf{P}} & \multicolumn{1}{c}{\textbf{R}} & \multicolumn{1}{c}{\textbf{F1}}\\ 
\hline 
\hline

\textbf{LR}  &	0.7570 &	0.7055	& 0.7409 &	0.7587 &	\underline{0.9129} &	0.8287 & 0.7513    & 0.4754  &   0.5823 \\
\textbf{MNB} &	0.7620 &	0.6960 &	0.7370 & 	0.7476 &	\textbf{0.9523} &	0.8376 & \textbf{0.8280} &   0.4167    & 0.5544  \\ 
\textbf{RF}	& 0.7430	& 0.6981 &	0.7316 &	0.7606	 & 0.8766 &	0.8145 &  0.6923    & 0.5016  &    0.5817\\ 
\textbf{SVM}  & 	0.7699 &	0.7347 &	0.7627 &	0.7880 &	0.8799 &	0.8314 & 0.7093  &  0.5279  &  0.6053 \\
\cdashline{1-10}

\textbf{BERT}                             & 0.7714                                                 & 0.7443                             & 0.7685                           & 0.8063                                 & 0.8499                              & 0.8275           & 0.6972   &  0.6286  &  0.6611               \\
\textbf{RoBERTa}                          & 0.7838                                                 & 0.7572                             & 0.7805                           & 0.8128                                 & 0.8639              & 0.8376              & 0.7204   &  0.6381   &  0.6768            \\
\textbf{XLNet}                            & \underline{0.8018}                                                 & \underline{0.7778}                             & \underline{0.7992}                           & \textbf{0.8284}                                 & 0.8746                              & \underline{0.8508}       & 0.7447    & 0.6688 &   \textbf{0.7047}                   \\
\textbf{BERTweet}                         & 0.7894                                                 & 0.7681                             & 0.7885                           & 0.8063                                 & 0.8499                              & 0.8384         & 0.7105  &  \textbf{0.6857} &   0.6979                 \\ \cdashline{1-10}
\textbf{AI Rational}         & 0.7905                       & 0.7606                            & 0.7852                          & 0.8076   & 0.8866  &  0.8453             & 0.7490    & 0.6159   &  0.6760      \\
\textbf{PoliMi FlatEarthers} & 0.7579                                                 & 0.7292                             & 0.7548                           & 0.7964                                 & 0.8394                              & 0.8173                     & 0.6761   & 0.6095  &  0.6411     \\
\textbf{NLP\&IR@UNED}        & 0.7905   & 0.7674           & 0.7887                           & 0.8252                                 & 0.8569                              & 0.8408               & 0.7201   &  0.6698   & 0.6941           \\
\textbf{Fight for 4230}      & 0.7860                   & 0.7585        & 0.7822           & 0.8114                                 & 0.8709                           & 0.8401               & 0.7289    & 0.6317   & 0.6769           \\
\textbf{Accenture}   & 0.7939          & 0.7647                             & 0.7888                    & 0.8105  &  0.8883  &  0.8476 & 0.7538  &  0.6222   & 0.6817                    \\
\textbf{Team Alex}            & 0.7872                                                 & 0.7653                             & 0.7861                           & \underline{0.8276}                                 & 0.8464                              & 0.8369             & 0.7086   & \underline{0.6794}   & 0.6937             \\ \cdashline{1-10}

\textbf{\name} & \textbf{0.8063} & \textbf{0.7802} & \textbf{0.8022} & 0.8229  &  0.8918  & \textbf{0.8559} & \underline{0.7678}  &  0.6508  &  \underline{0.7045}\\
\hline
\hline
\end{tabular}}
\label{tab:coarse-grained-result}
\end{table*}

\section{Results and Analysis}
\label{sec:results-and-analysis}
\subsection{Performance Comparison}
Even for humans, let alone automated systems, it can be challenging to discern check-worthy claims due to the high degree of subjectivity. The obstacle is heightened in the case of web-based short texts, which often lack linguistic soundness. Our aggregated results on the test set for coarse and fine-grained configurations are summarised in Tables \ref{tab:coarse-grained-result} and \ref{tab:fine-grained-result}, respectively. Most evaluation metrics show that \name\ beats all baseline systems. 

As previously stated, we primarily employ macro and weighted F1 scores to assess the efficacy of various claim check-worthiness models. According to the third and fourth columns of Table \ref{tab:fine-grained-result}, our model outperforms all existing baseline systems, including the current state-of-the-art system, AI Rational. On our dataset, \name\ accounts for a +$0.38$\% boost in weighted F1-score over the second-best performer, XLNet. There is also a considerable gain in accuracy -- our model has an accuracy of $80.63$, which is +$0.56$\% higher than the second-best performer. A palpable gain of about $5-10$\% across all metrics is ascertained when we transit from the classical machine learning architectures to the transformer-based models of BERT, RoBERTa, etc. This, in turn, emphasizes the significance of employing contextual word embeddings and transformer-based architectures for the task at hand. To capture this, through \name, we aim to amalgamate contextual attributes using a fine-tuned BERT model and linguistic features via part-of-speech and dependency tags. We find that \name\ outperforms all other systems in shared tasks, demonstrating that our framework can reasonably integrate plausible rationales alongside coarse-grained check-worthiness labels. 

To compute the base performance on fine-grained labels, we compare it to several baseline systems. Each question ($L1$-$L6$) is treated as a separate task in baseline systems. Overall, the experimental results reveal that no single model performs consistently best across all six supplementary questions. 

\begin{table}[th]
\centering
\renewcommand*{\arraystretch}{1.2}

\caption{Weighted F1-scores for fine-grained claim check-worthiness labels ($L1$-$L6$). Values in bold represent the highest scores, while
underlined values indicate the second-highest scores for the given metric. $\Delta$ refers to  }
\resizebox{0.70\textwidth}{!}{

\begin{tabular}{lcccccc}
\hline
\hline
\textbf{Model} & \textbf{L1} & \textbf{L2} & \textbf{L3} & \textbf{L4} & \textbf{L5} & \textbf{L6} \\
\hline 
\hline

\textbf{LR}  &	0.7313 & 0.6816 & 0.7640 & 0.7362 & 0.6913 & 0.7296\\
\textbf{MNB} &	0.7183 & 0.6499 & 0.7624 & 0.7136 & 0.6968 & 0.7293 \\ 
\textbf{RF}	& 0.7109 & 0.6743 & 0.7689 & 0.6863 & 0.6606 & 0.7110 \\ 
\textbf{SVM}  & 0.7359 & 0.7010 & 0.7743 & 0.7277 & 0.6862 & 0.7247 \\
\cdashline{1-7}

\textbf{BERT}  & 0.7346 & 0.6933 & 0.7857 & 0.5379 & 0.7021 & 0.7208    \\
\textbf{RoBERTa}   & 0.7723 & \textbf{0.7855}  &  0.7984 &   \underline{0.7866} &  0.7497 &  0.7182\\
\textbf{XLNet}  & \underline{0.7782} & 0.7631 & \textbf{0.8050} & 0.7527 & 0.7444 & \underline{0.7501} \\
\textbf{BERTweet} & 0.7700 & 0.7684 & 0.7845 & 0.7740 & \underline{0.7556} & \textbf{0.7666} \\   
\cdashline{1-7}
\textbf{\name}  & \textbf{0.7858} & \underline{0.7733} & \underline{0.7985} & \textbf{0.8095} & \textbf{0.7658} & 0.7421 \\
\hline
\hline
\end{tabular}}
\label{tab:fine-grained-result}
\end{table}

\subsection{Error Analysis}

\begin{table*}[ht]
\centering
\caption{Error analysis of the check-worthy outputs of \name\ using our Twitter dataset, \dname. For comparison, we also present the predictions of the best-performing baseline system, XLNet}
\label{tab:error-analysis}
\resizebox{\textwidth}{!}{
\begin{tabular}{c|p{7cm}|c|c|c}
    \hline
    \hline
    & \multirow{2}{*}{\bf Claim} & \multirow{2}{*}{\bf Gold} & \multicolumn{2}{c}{\bf Prediction} \\ \cline{4-5}
    & & & \bf \name & \bf XLNet \\ 
    \hline
    \hline
   
    $x_{1}$ & The only way to get cured of the Coronavirus is to drink Coronas & check-worthy & \textcolor{red}{non-check-worthy} & \textcolor{red}{non-check-worthy} \\ 
     \hline
    
    $x_{2}$ &  if we all stop breathing the spread would stop & non-check-worthy & \textcolor{red}{check-worthy} & non-check-worthy \\ \hline

    $x_{3}$ & science has shown that hydroxychloroquine does not work & check-worthy  & \textcolor{red}{non-check-worthy} &   \textcolor{red}{non-check-worthy} \\ \hline
    
    $x_{4}$ & No single VIRUS can wipe out the entire human race & non-check-worthy & non-check-worthy & check-worthy\\ \hline
    
    $x_{5}$ & disinfectant if injected will cure Covid 19 & check-worthy & \textcolor{red}{non check-worthy} &  \textcolor{red}{non-check-worthy} \\ \hline
    
    $x_{6}$ & if you inject or ingest Dettol, Lysol, bleach, or any other disinfectant, you almost certainly will not die of \#COVID19 & check-worthy & check-worthy & \textcolor{red}{non-check-worthy}. \\ \hline
    \hline

\end{tabular}}
\end{table*}

Due to the highly subjective nature of claims and the folksiness of online social media platforms, detecting check-worthy claims over these platforms poses a difficult task. Table \ref{tab:error-analysis} shows a few randomly selected instances from the test dataset and their gold and output labels as predicted by name. To compare, we also examine predictions from the best-performing baseline, XLNet.

XLNet and \name\ struggle to identify check-worthy claims in some cases, but \name\ perform exceptionally well in most cases. In example $x_{1}$, we see that both systems misclassify the claim as non-check-worthy and require assistance in comprehending the context of Coronas (beer). It follows that the systems have difficulty dealing with real-world knowledge. We witness similar behavior in example $x_{4}$, where the ineffectiveness of hydroxychloroquine towards COVID-19 confuses the models. In example $x_{2}$, the phrase \textit{``spread would stop''} causes the system to pay attention to it and incorrectly predicts it as a check-worthy claim. Given that it discusses one of the life-threatening COVID-19 cures, the gold label for $x_{5}$ indicates that the claim is check-worthy. However, the rambling writing style confuses both systems and yields inaccurate predictions.

\section{Conclusion}
\label{sec:conclusion}
In this work, we investigated how to merge two lines of research: claim check-worthiness detection and determining rationales for the check-worthy classification. We noticed that check-worthiness detection is difficult due to the considerable subjectivity involved, as there are significant variations in how annotators define what is deemed check-worthy. We illustrated how using different determining factors can help to make better decisions about check-worthy labels. Furthermore, improved rationales may assist manual fact-checkers in prioritizing fact-checking more swiftly. The lack of a suitably comprehensive annotated dataset is one of the primary constraints of fine-grained claim check-worthiness detection in online social media platforms. As a result, we developed a large Twitter corpus of more than $5k$ manually annotated claims for fine-grained claim check-worthiness identification. We devised a unified architecture, \name\ that outperforms existing state-of-the-art systems and illustrated how the abetting factors could be used to comprehend a fine-grained claim check-worthy system better. The results bestowed that our suggested model outperformed the best-performing baselines by $\geq$ $0.31$\% in macro F1-score and $\geq$ $0.38$\% in weighted F1-score. We acknowledge that in this work, we have concentrated on English; hence, in future work, we will seek to construct fine-grained claim check-worthiness detection models for other languages, particularly low-resource languages. According to our findings, \name\ fared better for the majority than the minority. As a result, our future work would focus on expanding the dataset and boosting the performance of the minority class. Finally, we intend to broaden our endeavor to incorporate multimodality.



\bibliographystyle{elsarticle-harv} 






\bibliography{bib}

\begin{thebibliography}{68}
\expandafter\ifx\csname natexlab\endcsname\relax\def\natexlab#1{#1}\fi
\providecommand{\url}[1]{\texttt{#1}}
\providecommand{\href}[2]{#2}
\providecommand{\path}[1]{#1}
\providecommand{\DOIprefix}{doi:}
\providecommand{\ArXivprefix}{arXiv:}
\providecommand{\URLprefix}{URL: }
\providecommand{\Pubmedprefix}{pmid:}
\providecommand{\doi}[1]{\href{http://dx.doi.org/#1}{\path{#1}}}
\providecommand{\Pubmed}[1]{\href{pmid:#1}{\path{#1}}}
\providecommand{\bibinfo}[2]{#2}
\ifx\xfnm\relax \def\xfnm[#1]{\unskip,\space#1}\fi
\bibitem[{Agez et~al.(2018)Agez, Bosc, Lespagnol, Mothe and Petitcol}]{agez2018irit}
\bibinfo{author}{Agez, R.}, \bibinfo{author}{Bosc, C.}, \bibinfo{author}{Lespagnol, C.}, \bibinfo{author}{Mothe, J.}, \bibinfo{author}{Petitcol, N.}, \bibinfo{year}{2018}.
\newblock \bibinfo{title}{Irit at checkthat! 2018}, in: \bibinfo{booktitle}{CLEF (Working Notes)}.
\bibitem[{Agrestia et~al.(2022)Agrestia, Hashemianb and Carmanc}]{agrestia2022polimi}
\bibinfo{author}{Agrestia, S.}, \bibinfo{author}{Hashemianb, A.}, \bibinfo{author}{Carmanc, M.}, \bibinfo{year}{2022}.
\newblock \bibinfo{title}{Polimi-flatearthers at checkthat! 2022: Gpt-3 applied to claim detection}.
\newblock \bibinfo{journal}{CLEF (Working Notes)} .
\bibitem[{Alam et~al.(2021)Alam, Dalvi, Shaar, Durrani, Mubarak, Nikolov, Da~San~Martino, Abdelali, Sajjad, Darwish et~al.}]{alam2021fighting}
\bibinfo{author}{Alam, F.}, \bibinfo{author}{Dalvi, F.}, \bibinfo{author}{Shaar, S.}, \bibinfo{author}{Durrani, N.}, \bibinfo{author}{Mubarak, H.}, \bibinfo{author}{Nikolov, A.}, \bibinfo{author}{Da~San~Martino, G.}, \bibinfo{author}{Abdelali, A.}, \bibinfo{author}{Sajjad, H.}, \bibinfo{author}{Darwish, K.}, et~al., \bibinfo{year}{2021}.
\newblock \bibinfo{title}{Fighting the covid-19 infodemic in social media: A holistic perspective and a call to arms.}, in: \bibinfo{booktitle}{Proc. of ICWSM}, pp. \bibinfo{pages}{913--922}.
\bibitem[{Allcott and Gentzkow(2017)}]{allcott2017social}
\bibinfo{author}{Allcott, H.}, \bibinfo{author}{Gentzkow, M.}, \bibinfo{year}{2017}.
\newblock \bibinfo{title}{Social media and fake news in the 2016 election}.
\newblock \bibinfo{journal}{Journal of economic perspectives} \bibinfo{volume}{31}, \bibinfo{pages}{211--36}.
\bibitem[{Altun and Kutlu(2019)}]{altun2019tobb}
\bibinfo{author}{Altun, B.}, \bibinfo{author}{Kutlu, M.}, \bibinfo{year}{2019}.
\newblock \bibinfo{title}{Tobb-etu at clef 2019: Prioritizing claims based on check-worthiness}, in: \bibinfo{booktitle}{Proc. of CEUR Workshop}, \bibinfo{organization}{CEUR-WS}.
\bibitem[{Antoun et~al.(2020)Antoun, Baly and Hajj}]{antoun2020arabert}
\bibinfo{author}{Antoun, W.}, \bibinfo{author}{Baly, F.}, \bibinfo{author}{Hajj, H.}, \bibinfo{year}{2020}.
\newblock \bibinfo{title}{Arabert: Transformer-based model for arabic language understanding}.
\newblock \bibinfo{journal}{arXiv:2003.00104} .
\bibitem[{Atanasova et~al.(2019)Atanasova, Nakov, Karadzhov, Mohtarami and Da~San~Martino}]{atanasova2019overview}
\bibinfo{author}{Atanasova, P.}, \bibinfo{author}{Nakov, P.}, \bibinfo{author}{Karadzhov, G.}, \bibinfo{author}{Mohtarami, M.}, \bibinfo{author}{Da~San~Martino, G.}, \bibinfo{year}{2019}.
\newblock \bibinfo{title}{Overview of the clef-2019 checkthat! lab: Automatic identification and verification of claims. task 1: Check-worthiness.}
\newblock \bibinfo{journal}{CLEF (Working Notes)} .
\bibitem[{Augenstein et~al.(2019)Augenstein, Lioma, Wang, Lima, Hansen, Hansen and Simonsen}]{augenstein2019multifc}
\bibinfo{author}{Augenstein, I.}, \bibinfo{author}{Lioma, C.}, \bibinfo{author}{Wang, D.}, \bibinfo{author}{Lima, L.C.}, \bibinfo{author}{Hansen, C.}, \bibinfo{author}{Hansen, C.}, \bibinfo{author}{Simonsen, J.G.}, \bibinfo{year}{2019}.
\newblock \bibinfo{title}{Multifc: A real-world multi-domain dataset for evidence-based fact checking of claims}, in: \bibinfo{booktitle}{Proc. of EMNLP-IJCNLP}, pp. \bibinfo{pages}{4685--4697}.
\bibitem[{Barr{\'o}n-Cede{\~n}o et~al.(2020)Barr{\'o}n-Cede{\~n}o, Elsayed, Nakov, Da~San~Martino, Hasanain, Suwaileh, Haouari, Babulkov, Hamdan, Nikolov et~al.}]{barron2020overview}
\bibinfo{author}{Barr{\'o}n-Cede{\~n}o, A.}, \bibinfo{author}{Elsayed, T.}, \bibinfo{author}{Nakov, P.}, \bibinfo{author}{Da~San~Martino, G.}, \bibinfo{author}{Hasanain, M.}, \bibinfo{author}{Suwaileh, R.}, \bibinfo{author}{Haouari, F.}, \bibinfo{author}{Babulkov, N.}, \bibinfo{author}{Hamdan, B.}, \bibinfo{author}{Nikolov, A.}, et~al., \bibinfo{year}{2020}.
\newblock \bibinfo{title}{Overview of checkthat! 2020: Automatic identification and verification of claims in social media}, in: \bibinfo{booktitle}{CLEF 2020}, \bibinfo{organization}{Springer}. pp. \bibinfo{pages}{215--236}.
\bibitem[{Barron-Cedeno et~al.(2020)Barron-Cedeno, Elsayed, Nakov, Martino, Hasanain, Suwaileh, Haouari, Babulkov, Hamdan, Nikolov, Shaar and Ali}]{barroncedeno2020overview}
\bibinfo{author}{Barron-Cedeno, A.}, \bibinfo{author}{Elsayed, T.}, \bibinfo{author}{Nakov, P.}, \bibinfo{author}{Martino, G.D.S.}, \bibinfo{author}{Hasanain, M.}, \bibinfo{author}{Suwaileh, R.}, \bibinfo{author}{Haouari, F.}, \bibinfo{author}{Babulkov, N.}, \bibinfo{author}{Hamdan, B.}, \bibinfo{author}{Nikolov, A.}, \bibinfo{author}{Shaar, S.}, \bibinfo{author}{Ali, Z.S.}, \bibinfo{year}{2020}.
\newblock \bibinfo{title}{Overview of checkthat! 2020: Automatic identification and verification of claims in social media} \href{http://arxiv.org/abs/2007.07997}{{\tt arXiv:2007.07997}}.
\bibitem[{Barr{\'o}n-Cedeno et~al.(2018)Barr{\'o}n-Cedeno, Elsayed, Suwaileh, M{\`a}rquez, Atanasova, Zaghouani, Kyuchukov, Da~San~Martino and Nakov}]{barron2018overview}
\bibinfo{author}{Barr{\'o}n-Cedeno, A.}, \bibinfo{author}{Elsayed, T.}, \bibinfo{author}{Suwaileh, R.}, \bibinfo{author}{M{\`a}rquez, L.}, \bibinfo{author}{Atanasova, P.}, \bibinfo{author}{Zaghouani, W.}, \bibinfo{author}{Kyuchukov, S.}, \bibinfo{author}{Da~San~Martino, G.}, \bibinfo{author}{Nakov, P.}, \bibinfo{year}{2018}.
\newblock \bibinfo{title}{Overview of the clef-2018 checkthat! lab on automatic identification and verification of political claims. task 2: Factuality.}
\newblock \bibinfo{journal}{CLEF (Working Notes)} \bibinfo{volume}{2125}.
\bibitem[{Beech(2020)}]{beech_2020}
\bibinfo{author}{Beech, M.}, \bibinfo{year}{2020}.
\newblock \bibinfo{title}{Covid-19 pushes up internet use 70\% and streaming more than 12\%, first figures reveal}.
\newblock \URLprefix \url{https://www.forbes.com/sites/markbeech/2020/03/25/covid-19-pushes-up-internet-use-70-streaming-more-than-12-first- \\figures-reveal}.
\bibitem[{Brown et~al.(2020)Brown, Mann, Ryder, Subbiah, Kaplan, Dhariwal, Neelakantan, Shyam, Sastry, Askell et~al.}]{brown2020language}
\bibinfo{author}{Brown, T.}, \bibinfo{author}{Mann, B.}, \bibinfo{author}{Ryder, N.}, \bibinfo{author}{Subbiah, M.}, \bibinfo{author}{Kaplan, J.D.}, \bibinfo{author}{Dhariwal, P.}, \bibinfo{author}{Neelakantan, A.}, \bibinfo{author}{Shyam, P.}, \bibinfo{author}{Sastry, G.}, \bibinfo{author}{Askell, A.}, et~al., \bibinfo{year}{2020}.
\newblock \bibinfo{title}{Language models are few-shot learners}.
\newblock \bibinfo{journal}{NeurIPS} \bibinfo{volume}{33}, \bibinfo{pages}{1877--1901}.
\bibitem[{Cheema et~al.(2020)Cheema, Hakimov and Ewerth}]{cheema2020check_square}
\bibinfo{author}{Cheema, G.S.}, \bibinfo{author}{Hakimov, S.}, \bibinfo{author}{Ewerth, R.}, \bibinfo{year}{2020}.
\newblock \bibinfo{title}{Check\_square at checkthat! 2020: Claim detection in social media via fusion of transformer and syntactic features}.
\newblock \bibinfo{journal}{arXiv:2007.10534} .
\bibitem[{Cherubini and Graves()}]{cherubini_graves}
\bibinfo{author}{Cherubini, F.}, \bibinfo{author}{Graves, L.}, .
\newblock \bibinfo{title}{The rise of fact-checking sites in europe}.
\newblock \URLprefix \url{https://reutersinstitute.politics.ox.ac.uk/our-research/rise-fact-checking-sites-europe}.
\bibitem[{Coca et~al.(2019)Coca, Cusmuliuc and Iftene}]{coca2019checkthat}
\bibinfo{author}{Coca, L.G.}, \bibinfo{author}{Cusmuliuc, C.G.}, \bibinfo{author}{Iftene, A.}, \bibinfo{year}{2019}.
\newblock \bibinfo{title}{Checkthat! 2019 uaics.}, in: \bibinfo{booktitle}{CLEF (Working Notes)}.
\bibitem[{Cohen(1960)}]{cohen1960}
\bibinfo{author}{Cohen, J.}, \bibinfo{year}{1960}.
\newblock \bibinfo{title}{{A Coefficient of Agreement for Nominal Scales}}.
\newblock \bibinfo{journal}{Educational and Psychological Measurement} \bibinfo{volume}{20}, \bibinfo{pages}{37}.
\bibitem[{Cusmuliuc et~al.(2020)Cusmuliuc, Coca and Iftene}]{cusmuliuc2020uaics}
\bibinfo{author}{Cusmuliuc, C.G.}, \bibinfo{author}{Coca, L.G.}, \bibinfo{author}{Iftene, A.}, \bibinfo{year}{2020}.
\newblock \bibinfo{title}{Uaics at checkthat! 2020 fact-checking claim prioritization.}, in: \bibinfo{booktitle}{CLEF (Working Notes)}.
\bibitem[{Das et~al.(2023)Das, Liu, Kovatchev and Lease}]{das2023state}
\bibinfo{author}{Das, A.}, \bibinfo{author}{Liu, H.}, \bibinfo{author}{Kovatchev, V.}, \bibinfo{author}{Lease, M.}, \bibinfo{year}{2023}.
\newblock \bibinfo{title}{The state of human-centered nlp technology for fact-checking}.
\newblock \bibinfo{journal}{Information processing \& management} \bibinfo{volume}{60}, \bibinfo{pages}{103219}.
\bibitem[{Devlin et~al.(2019)Devlin, Chang, Lee and Toutanova}]{devlin2018bert}
\bibinfo{author}{Devlin, J.}, \bibinfo{author}{Chang, M.W.}, \bibinfo{author}{Lee, K.}, \bibinfo{author}{Toutanova, K.}, \bibinfo{year}{2019}.
\newblock \bibinfo{title}{{BERT}: Pre-training of deep bidirectional transformers for language understanding}, in: \bibinfo{booktitle}{Proc. of NAACL-HLT}, \bibinfo{publisher}{ACL}. pp. \bibinfo{pages}{4171--4186}.
\bibitem[{Dhar et~al.(2019)Dhar, Dutta and Das}]{dhar2019hybrid}
\bibinfo{author}{Dhar, R.}, \bibinfo{author}{Dutta, S.}, \bibinfo{author}{Das, D.}, \bibinfo{year}{2019}.
\newblock \bibinfo{title}{A hybrid model to rank sentences for check-worthiness.}, in: \bibinfo{booktitle}{CLEF (Working Notes)}.
\bibitem[{Favano et~al.(2019)Favano, Carman and Lanzi}]{favano2019theearthisflat}
\bibinfo{author}{Favano, L.}, \bibinfo{author}{Carman, M.J.}, \bibinfo{author}{Lanzi, P.L.}, \bibinfo{year}{2019}.
\newblock \bibinfo{title}{Theearthisflat's submission to clef'19checkthat! challenge}, in: \bibinfo{booktitle}{CLEF (Working Notes)}, \bibinfo{organization}{CEUR-WS. org}. pp. \bibinfo{pages}{1--11}.
\bibitem[{Gencheva et~al.(2017)Gencheva, Nakov, M{\`a}rquez, Barr{\'o}n-Cede{\~n}o and Koychev}]{gencheva2017context}
\bibinfo{author}{Gencheva, P.}, \bibinfo{author}{Nakov, P.}, \bibinfo{author}{M{\`a}rquez, L.}, \bibinfo{author}{Barr{\'o}n-Cede{\~n}o, A.}, \bibinfo{author}{Koychev, I.}, \bibinfo{year}{2017}.
\newblock \bibinfo{title}{A context-aware approach for detecting worth-checking claims in political debates}, in: \bibinfo{booktitle}{Proc. of RANLP}, pp. \bibinfo{pages}{267--276}.
\bibitem[{Ghanem et~al.(2018)Ghanem, G{\'o}mez, Rangel and Rosso}]{ghanem2018upv}
\bibinfo{author}{Ghanem, B.}, \bibinfo{author}{G{\'o}mez, M.M.y.}, \bibinfo{author}{Rangel, F.}, \bibinfo{author}{Rosso, P.}, \bibinfo{year}{2018}.
\newblock \bibinfo{title}{Upv-inaoe-autoritas-check that: Preliminary approach for checking worthiness of claims}, in: \bibinfo{booktitle}{CLEF (Working Notes)}, pp. \bibinfo{pages}{1--6}.
\bibitem[{Grave et~al.(2018)Grave, Bojanowski, Gupta, Joulin and Mikolov}]{grave2018learning}
\bibinfo{author}{Grave, E.}, \bibinfo{author}{Bojanowski, P.}, \bibinfo{author}{Gupta, P.}, \bibinfo{author}{Joulin, A.}, \bibinfo{author}{Mikolov, T.}, \bibinfo{year}{2018}.
\newblock \bibinfo{title}{Learning word vectors for 157 languages}.
\newblock \bibinfo{journal}{preprint arXiv:1802.06893} .
\bibitem[{Gupta et~al.(2021)Gupta, Singh, Sundriyal, Akhtar and Chakraborty}]{gupta2021lesa}
\bibinfo{author}{Gupta, S.}, \bibinfo{author}{Singh, P.}, \bibinfo{author}{Sundriyal, M.}, \bibinfo{author}{Akhtar, M.S.}, \bibinfo{author}{Chakraborty, T.}, \bibinfo{year}{2021}.
\newblock \bibinfo{title}{Lesa: Linguistic encapsulation and semantic amalgamation based generalised claim detection from online content}, in: \bibinfo{booktitle}{Proc. of EACL}, pp. \bibinfo{pages}{3178--3188}.
\bibitem[{Hansen et~al.(2019)Hansen, Hansen, Simonsen and Lioma}]{hansen2019neural}
\bibinfo{author}{Hansen, C.}, \bibinfo{author}{Hansen, C.}, \bibinfo{author}{Simonsen, J.}, \bibinfo{author}{Lioma, C.}, \bibinfo{year}{2019}.
\newblock \bibinfo{title}{Neural weakly supervised fact check-worthiness detection with contrastive sampling-based ranking loss}, in: \bibinfo{booktitle}{CLEF (Working Notes)}.
\bibitem[{Hansen et~al.(2018)Hansen, Hansen, Simonsen and Lioma}]{hansen2018copenhagen}
\bibinfo{author}{Hansen, C.}, \bibinfo{author}{Hansen, C.}, \bibinfo{author}{Simonsen, J.G.}, \bibinfo{author}{Lioma, C.}, \bibinfo{year}{2018}.
\newblock \bibinfo{title}{The copenhagen team participation in the check-worthiness task of the competition of automatic identification and verification of claims in political debates of the clef-2018 checkthat! lab.}, in: \bibinfo{booktitle}{CLEF (Working Notes)}.
\bibitem[{Hasanain and Elsayed(2020)}]{hasanain2020bigir}
\bibinfo{author}{Hasanain, M.}, \bibinfo{author}{Elsayed, T.}, \bibinfo{year}{2020}.
\newblock \bibinfo{title}{bigir at checkthat! 2020: Multilingual bert for ranking arabic tweets by check-worthiness.}, in: \bibinfo{booktitle}{CLEF (Working Notes)}.
\bibitem[{Hassan et~al.(2015)Hassan, Li and Tremayne}]{hassan2015detecting}
\bibinfo{author}{Hassan, N.}, \bibinfo{author}{Li, C.}, \bibinfo{author}{Tremayne, M.}, \bibinfo{year}{2015}.
\newblock \bibinfo{title}{Detecting check-worthy factual claims in presidential debates}, in: \bibinfo{booktitle}{Proc. of CIKM}, pp. \bibinfo{pages}{1835--1838}.
\bibitem[{Hassan et~al.(2017)Hassan, Zhang, Arslan, Caraballo, Jimenez, Gawsane, Hasan, Joseph, Kulkarni, Nayak et~al.}]{hassan2017claimbuster}
\bibinfo{author}{Hassan, N.}, \bibinfo{author}{Zhang, G.}, \bibinfo{author}{Arslan, F.}, \bibinfo{author}{Caraballo, J.}, \bibinfo{author}{Jimenez, D.}, \bibinfo{author}{Gawsane, S.}, \bibinfo{author}{Hasan, S.}, \bibinfo{author}{Joseph, M.}, \bibinfo{author}{Kulkarni, A.}, \bibinfo{author}{Nayak, A.K.}, et~al., \bibinfo{year}{2017}.
\newblock \bibinfo{title}{Claimbuster: The first-ever end-to-end fact-checking system}.
\newblock \bibinfo{journal}{Proc. of the VLDB Endowment 2017} \bibinfo{volume}{10}, \bibinfo{pages}{1945--1948}.
\bibitem[{Hoang and Mothe(2018)}]{hoang2018location}
\bibinfo{author}{Hoang, T.B.N.}, \bibinfo{author}{Mothe, J.}, \bibinfo{year}{2018}.
\newblock \bibinfo{title}{Location extraction from tweets}.
\newblock \bibinfo{journal}{Information Processing \& Management} \bibinfo{volume}{54}, \bibinfo{pages}{129--144}.
\bibitem[{Jaradat et~al.(2018)Jaradat, Gencheva, Barr{\'o}n-Cede{\~n}o, M{\`a}rquez and Nakov}]{jaradat2018claimrank}
\bibinfo{author}{Jaradat, I.}, \bibinfo{author}{Gencheva, P.}, \bibinfo{author}{Barr{\'o}n-Cede{\~n}o, A.}, \bibinfo{author}{M{\`a}rquez, L.}, \bibinfo{author}{Nakov, P.}, \bibinfo{year}{2018}.
\newblock \bibinfo{title}{Claimrank: Detecting check-worthy claims in arabic and english}, in: \bibinfo{booktitle}{Proc. of NAACL: Demonstrations}, pp. \bibinfo{pages}{26--30}.
\bibitem[{Kartal and Kutlu(2020)}]{kartal2020tobb}
\bibinfo{author}{Kartal, Y.S.}, \bibinfo{author}{Kutlu, M.}, \bibinfo{year}{2020}.
\newblock \bibinfo{title}{Tobb etu at checkthat! 2020: Prioritizing english and arabic claims based on check-worthiness.}, in: \bibinfo{booktitle}{CLEF (Working Notes)}.
\bibitem[{Kemp(2022)}]{kemp_2022}
\bibinfo{author}{Kemp, S.}, \bibinfo{year}{2022}.
\newblock \bibinfo{title}{Digital 2022: Global overview report - datareportal – global digital insights}.
\newblock \URLprefix \url{https://datareportal.com/reports/digital-2022-global-overview-report}.
\bibitem[{Konstantinovskiy et~al.(2021)Konstantinovskiy, Price, Babakar and Zubiaga}]{konstantinovskiy2021toward}
\bibinfo{author}{Konstantinovskiy, L.}, \bibinfo{author}{Price, O.}, \bibinfo{author}{Babakar, M.}, \bibinfo{author}{Zubiaga, A.}, \bibinfo{year}{2021}.
\newblock \bibinfo{title}{Toward automated factchecking: Developing an annotation schema and benchmark for consistent automated claim detection}.
\newblock \bibinfo{journal}{Digital threats: research and practice} \bibinfo{volume}{2}, \bibinfo{pages}{1--16}.
\bibitem[{Lazer et~al.(2018)Lazer, Baum, Benkler, Berinsky, Greenhill, Menczer, Metzger, Nyhan, Pennycook, Rothschild et~al.}]{lazer2018science}
\bibinfo{author}{Lazer, D.M.}, \bibinfo{author}{Baum, M.A.}, \bibinfo{author}{Benkler, Y.}, \bibinfo{author}{Berinsky, A.J.}, \bibinfo{author}{Greenhill, K.M.}, \bibinfo{author}{Menczer, F.}, \bibinfo{author}{Metzger, M.J.}, \bibinfo{author}{Nyhan, B.}, \bibinfo{author}{Pennycook, G.}, \bibinfo{author}{Rothschild, D.}, et~al., \bibinfo{year}{2018}.
\newblock \bibinfo{title}{The science of fake news}.
\newblock \bibinfo{journal}{Science} \bibinfo{volume}{359}, \bibinfo{pages}{1094--1096}.
\bibitem[{Levy et~al.(2014)Levy, Bilu, Hershcovich, Aharoni and Slonim}]{levy2014context}
\bibinfo{author}{Levy, R.}, \bibinfo{author}{Bilu, Y.}, \bibinfo{author}{Hershcovich, D.}, \bibinfo{author}{Aharoni, E.}, \bibinfo{author}{Slonim, N.}, \bibinfo{year}{2014}.
\newblock \bibinfo{title}{Context dependent claim detection}, in: \bibinfo{booktitle}{Proc. of COLING}, pp. \bibinfo{pages}{1489--1500}.
\bibitem[{Levy et~al.(2017)Levy, Gretz, Sznajder, Hummel, Aharonov and Slonim}]{levy2017unsupervised}
\bibinfo{author}{Levy, R.}, \bibinfo{author}{Gretz, S.}, \bibinfo{author}{Sznajder, B.}, \bibinfo{author}{Hummel, S.}, \bibinfo{author}{Aharonov, R.}, \bibinfo{author}{Slonim, N.}, \bibinfo{year}{2017}.
\newblock \bibinfo{title}{Unsupervised corpus--wide claim detection}, in: \bibinfo{booktitle}{Proc. of Workshop on Argument Mining}, pp. \bibinfo{pages}{79--84}.
\bibitem[{Liu et~al.(2019)Liu, Ott, Goyal, Du, Joshi, Chen, Levy, Lewis, Zettlemoyer and Stoyanov}]{liu2019roberta}
\bibinfo{author}{Liu, Y.}, \bibinfo{author}{Ott, M.}, \bibinfo{author}{Goyal, N.}, \bibinfo{author}{Du, J.}, \bibinfo{author}{Joshi, M.}, \bibinfo{author}{Chen, D.}, \bibinfo{author}{Levy, O.}, \bibinfo{author}{Lewis, M.}, \bibinfo{author}{Zettlemoyer, L.}, \bibinfo{author}{Stoyanov, V.}, \bibinfo{year}{2019}.
\newblock \bibinfo{title}{Roberta: A robustly optimized bert pretraining approach}.
\newblock \bibinfo{journal}{preprint arXiv:1907.11692} .
\bibitem[{Martinez-Rico et~al.(2021)Martinez-Rico, Mart{\'\i}nez-Romo and Araujo}]{martinez2021nlp}
\bibinfo{author}{Martinez-Rico, J.R.}, \bibinfo{author}{Mart{\'\i}nez-Romo, J.}, \bibinfo{author}{Araujo, L.}, \bibinfo{year}{2021}.
\newblock \bibinfo{title}{Nlp\&ir@ uned at checkthat! 2021: Check-worthiness estimation and fake news detection using transformer models.}, in: \bibinfo{booktitle}{CLEF (Working Notes)}, pp. \bibinfo{pages}{545--557}.
\bibitem[{Mikolov et~al.(2013)Mikolov, Chen, Corrado and Dean}]{mikolov2013efficient}
\bibinfo{author}{Mikolov, T.}, \bibinfo{author}{Chen, K.}, \bibinfo{author}{Corrado, G.}, \bibinfo{author}{Dean, J.}, \bibinfo{year}{2013}.
\newblock \bibinfo{title}{Efficient estimation of word representations in vector space}.
\newblock \bibinfo{journal}{arXiv: 1301.3781} \href{http://arxiv.org/abs/1301.3781}{{\tt arXiv:1301.3781}}.
\bibitem[{Nakov et~al.(2022)Nakov, Barr{\'o}n-Cede{\~n}o, da~San~Martino, Alam, Stru{\ss}, Mandl, M{\'\i}guez, Caselli, Kutlu, Zaghouani et~al.}]{nakov2022overview}
\bibinfo{author}{Nakov, P.}, \bibinfo{author}{Barr{\'o}n-Cede{\~n}o, A.}, \bibinfo{author}{da~San~Martino, G.}, \bibinfo{author}{Alam, F.}, \bibinfo{author}{Stru{\ss}, J.M.}, \bibinfo{author}{Mandl, T.}, \bibinfo{author}{M{\'\i}guez, R.}, \bibinfo{author}{Caselli, T.}, \bibinfo{author}{Kutlu, M.}, \bibinfo{author}{Zaghouani, W.}, et~al., \bibinfo{year}{2022}.
\newblock \bibinfo{title}{Overview of the clef--2022 checkthat! lab on fighting the covid-19 infodemic and fake news detection}, in: \bibinfo{booktitle}{CLEF (working notes)}, \bibinfo{organization}{Springer}. pp. \bibinfo{pages}{495--520}.
\bibitem[{Nakov et~al.(2021)Nakov, Da~San~Martino, Elsayed, Barr{\'o}n-Cede{\~n}o, M{\'\i}guez, Shaar, Alam, Haouari, Hasanain, Mansour et~al.}]{nakov2021overview}
\bibinfo{author}{Nakov, P.}, \bibinfo{author}{Da~San~Martino, G.}, \bibinfo{author}{Elsayed, T.}, \bibinfo{author}{Barr{\'o}n-Cede{\~n}o, A.}, \bibinfo{author}{M{\'\i}guez, R.}, \bibinfo{author}{Shaar, S.}, \bibinfo{author}{Alam, F.}, \bibinfo{author}{Haouari, F.}, \bibinfo{author}{Hasanain, M.}, \bibinfo{author}{Mansour, W.}, et~al., \bibinfo{year}{2021}.
\newblock \bibinfo{title}{Overview of the clef--2021 checkthat! lab on detecting check-worthy claims, previously fact-checked claims, and fake news}, in: \bibinfo{booktitle}{CLEF (Working Notes)}, \bibinfo{organization}{Springer}. pp. \bibinfo{pages}{264--291}.
\bibitem[{Nguyen et~al.(2020)Nguyen, Vu and Nguyen}]{nguyen2020bertweet}
\bibinfo{author}{Nguyen, D.Q.}, \bibinfo{author}{Vu, T.}, \bibinfo{author}{Nguyen, A.T.}, \bibinfo{year}{2020}.
\newblock \bibinfo{title}{Bertweet: A pre-trained language model for english tweets}, in: \bibinfo{booktitle}{Proc. of EMNLP: System Demonstrations}, pp. \bibinfo{pages}{9--14}.
\bibitem[{Nikolov et~al.(2020)Nikolov, Martino, Koychev and Nakov}]{nikolov2020team}
\bibinfo{author}{Nikolov, A.}, \bibinfo{author}{Martino, G.D.S.}, \bibinfo{author}{Koychev, I.}, \bibinfo{author}{Nakov, P.}, \bibinfo{year}{2020}.
\newblock \bibinfo{title}{Team alex at clef checkthat! 2020: Identifying check-worthy tweets with transformer models}.
\newblock \bibinfo{journal}{arXiv:2009.02931} \href{http://arxiv.org/abs/2009.02931}{{\tt arXiv:2009.02931}}.
\bibitem[{Patwari et~al.(2017)Patwari, Goldwasser and Bagchi}]{patwari2017tathya}
\bibinfo{author}{Patwari, A.}, \bibinfo{author}{Goldwasser, D.}, \bibinfo{author}{Bagchi, S.}, \bibinfo{year}{2017}.
\newblock \bibinfo{title}{Tathya: A multi-classifier system for detecting check-worthy statements in political debates}, in: \bibinfo{booktitle}{Proc. of CIKM}, pp. \bibinfo{pages}{2259--2262}.
\bibitem[{Peldszus and Stede(2015)}]{peldszus-stede-2015-joint}
\bibinfo{author}{Peldszus, A.}, \bibinfo{author}{Stede, M.}, \bibinfo{year}{2015}.
\newblock \bibinfo{title}{Joint prediction in {MST}-style discourse parsing for argumentation mining}, in: \bibinfo{booktitle}{Proc. of EMNLP}, \bibinfo{publisher}{ACL}, \bibinfo{address}{Lisbon, Portugal}. pp. \bibinfo{pages}{938--948}.
\bibitem[{Ratkiewicz et~al.(2011)Ratkiewicz, Conover, Meiss, Gon{\c{c}}alves, Flammini and Menczer}]{ratkiewicz2011detecting}
\bibinfo{author}{Ratkiewicz, J.}, \bibinfo{author}{Conover, M.}, \bibinfo{author}{Meiss, M.}, \bibinfo{author}{Gon{\c{c}}alves, B.}, \bibinfo{author}{Flammini, A.}, \bibinfo{author}{Menczer, F.}, \bibinfo{year}{2011}.
\newblock \bibinfo{title}{Detecting and tracking political abuse in social media}, in: \bibinfo{booktitle}{Proc. of ICWSM}, pp. \bibinfo{pages}{297--304}.
\bibitem[{Sanh et~al.(2019)Sanh, Debut, Chaumond and Wolf}]{sanh2019distilbert}
\bibinfo{author}{Sanh, V.}, \bibinfo{author}{Debut, L.}, \bibinfo{author}{Chaumond, J.}, \bibinfo{author}{Wolf, T.}, \bibinfo{year}{2019}.
\newblock \bibinfo{title}{Distilbert, a distilled version of bert: smaller, faster, cheaper and lighter}.
\newblock \bibinfo{journal}{arXiv:1910.01108} .
\bibitem[{Savchev(2022)}]{savchev2022ai}
\bibinfo{author}{Savchev, A.}, \bibinfo{year}{2022}.
\newblock \bibinfo{title}{Ai rational at checkthat! 2022: using transformer models for tweet classification}.
\newblock \bibinfo{journal}{CLEF (Working Notes)} .
\bibitem[{Sefidbakht et~al.(2020)Sefidbakht, Lotfi, Jalli, Moghadami, Sabetian and Iranpour}]{sefidbakht2020methanol}
\bibinfo{author}{Sefidbakht, S.}, \bibinfo{author}{Lotfi, M.}, \bibinfo{author}{Jalli, R.}, \bibinfo{author}{Moghadami, M.}, \bibinfo{author}{Sabetian, G.}, \bibinfo{author}{Iranpour, P.}, \bibinfo{year}{2020}.
\newblock \bibinfo{title}{Methanol toxicity outbreak: when fear of covid-19 goes viral}.
\newblock \bibinfo{journal}{Emergency medicine journal} \bibinfo{volume}{37}, \bibinfo{pages}{416--416}.
\bibitem[{Shaar et~al.(2021)Shaar, Hasanain, Hamdan, Ali, Haouari, Nikolov, Kutlu, Kartal, Alam, Da~San~Martino et~al.}]{shaar2021overview}
\bibinfo{author}{Shaar, S.}, \bibinfo{author}{Hasanain, M.}, \bibinfo{author}{Hamdan, B.}, \bibinfo{author}{Ali, Z.S.}, \bibinfo{author}{Haouari, F.}, \bibinfo{author}{Nikolov, A.}, \bibinfo{author}{Kutlu, M.}, \bibinfo{author}{Kartal, Y.S.}, \bibinfo{author}{Alam, F.}, \bibinfo{author}{Da~San~Martino, G.}, et~al., \bibinfo{year}{2021}.
\newblock \bibinfo{title}{Overview of the clef-2021 checkthat! lab task 1 on check-worthiness estimation in tweets and political debates}, in: \bibinfo{booktitle}{CLEF (Working Notes)}.
\bibitem[{Stab and Gurevych(2017)}]{stab-gurevych-2017-parsing}
\bibinfo{author}{Stab, C.}, \bibinfo{author}{Gurevych, I.}, \bibinfo{year}{2017}.
\newblock \bibinfo{title}{Parsing argumentation structures in persuasive essays}.
\newblock \bibinfo{journal}{Computational Linguistics} \bibinfo{volume}{43}, \bibinfo{pages}{619--659}.
\bibitem[{Sundriyal et~al.(2022a)Sundriyal, Kulkarni, Pulastya, Akhtar and Chakraborty}]{sundriyal2022empowering}
\bibinfo{author}{Sundriyal, M.}, \bibinfo{author}{Kulkarni, A.}, \bibinfo{author}{Pulastya, V.}, \bibinfo{author}{Akhtar, M.S.}, \bibinfo{author}{Chakraborty, T.}, \bibinfo{year}{2022}a.
\newblock \bibinfo{title}{Empowering the fact-checkers! automatic identification of claim spans on twitter}.
\newblock \bibinfo{journal}{arXiv:2210.04710} .
\bibitem[{Sundriyal et~al.(2022b)Sundriyal, Malhotra, Akhtar, Sengupta, Fano and Chakraborty}]{sundriyal-etal-2022-document}
\bibinfo{author}{Sundriyal, M.}, \bibinfo{author}{Malhotra, G.}, \bibinfo{author}{Akhtar, M.S.}, \bibinfo{author}{Sengupta, S.}, \bibinfo{author}{Fano, A.}, \bibinfo{author}{Chakraborty, T.}, \bibinfo{year}{2022}b.
\newblock \bibinfo{title}{Document retrieval and claim verification to mitigate {COVID}-19 misinformation}, in: \bibinfo{booktitle}{Proc. of workshop on CONSTRAINT}, \bibinfo{publisher}{ACL}. pp. \bibinfo{pages}{66--74}.
\bibitem[{Sundriyal et~al.(2021)Sundriyal, Singh, Akhtar, Sengupta and Chakraborty}]{sundriyal2021desyr}
\bibinfo{author}{Sundriyal, M.}, \bibinfo{author}{Singh, P.}, \bibinfo{author}{Akhtar, M.S.}, \bibinfo{author}{Sengupta, S.}, \bibinfo{author}{Chakraborty, T.}, \bibinfo{year}{2021}.
\newblock \bibinfo{title}{Desyr: definition and syntactic representation based claim detection on the web}, in: \bibinfo{booktitle}{Proc. of CIKM}, pp. \bibinfo{pages}{1764--1773}.
\bibitem[{Thorne et~al.(2018a)Thorne, Vlachos, Christodoulopoulos and Mittal}]{thorne2018fever}
\bibinfo{author}{Thorne, J.}, \bibinfo{author}{Vlachos, A.}, \bibinfo{author}{Christodoulopoulos, C.}, \bibinfo{author}{Mittal, A.}, \bibinfo{year}{2018}a.
\newblock \bibinfo{title}{Fever: a large-scale dataset for fact extraction and verification}.
\newblock \bibinfo{journal}{arXiv:1803.05355} .
\bibitem[{Thorne et~al.(2018b)Thorne, Vlachos, Cocarascu, Christodoulopoulos and Mittal}]{thorne-etal-2018-fact}
\bibinfo{author}{Thorne, J.}, \bibinfo{author}{Vlachos, A.}, \bibinfo{author}{Cocarascu, O.}, \bibinfo{author}{Christodoulopoulos, C.}, \bibinfo{author}{Mittal, A.}, \bibinfo{year}{2018}b.
\newblock \bibinfo{title}{The fact extraction and {VER}ification ({FEVER}) shared task}, in: \bibinfo{booktitle}{Proc. of Workshop on FEVER}, \bibinfo{publisher}{ACL}, \bibinfo{address}{Brussels, Belgium}. pp. \bibinfo{pages}{1--9}.
\bibitem[{Toraman et~al.(2022)Toraman, Ozcelik, {\c{S}}ahinu{\c{c}} and Sahin}]{toraman2022arc}
\bibinfo{author}{Toraman, C.}, \bibinfo{author}{Ozcelik, O.}, \bibinfo{author}{{\c{S}}ahinu{\c{c}}, F.}, \bibinfo{author}{Sahin, U.}, \bibinfo{year}{2022}.
\newblock \bibinfo{title}{Arc-nlp at checkthat! 2022: contradiction for harmful tweet detection}.
\newblock \bibinfo{journal}{CLEF (Working Notes)} .
\bibitem[{Toulmin(2003)}]{toulmin2003uses}
\bibinfo{author}{Toulmin, S.E.}, \bibinfo{year}{2003}.
\newblock \bibinfo{title}{The uses of argument}.
\newblock \bibinfo{publisher}{Cambridge university press}.
\bibitem[{Vasileva et~al.(2019)Vasileva, Atanasova, M{\`a}rquez, Barr{\'o}n-Cede{\~n}o and Nakov}]{vasileva2019takes}
\bibinfo{author}{Vasileva, S.}, \bibinfo{author}{Atanasova, P.}, \bibinfo{author}{M{\`a}rquez, L.}, \bibinfo{author}{Barr{\'o}n-Cede{\~n}o, A.}, \bibinfo{author}{Nakov, P.}, \bibinfo{year}{2019}.
\newblock \bibinfo{title}{It takes nine to smell a rat: Neural multi-task learning for check-worthiness prediction}, in: \bibinfo{booktitle}{Proc. of RANLP}, pp. \bibinfo{pages}{1229--1239}.
\bibitem[{Vlachos and Riedel(2014)}]{vlachos-riedel-2014-fact}
\bibinfo{author}{Vlachos, A.}, \bibinfo{author}{Riedel, S.}, \bibinfo{year}{2014}.
\newblock \bibinfo{title}{Fact checking: Task definition and dataset construction}, in: \bibinfo{booktitle}{Proc. of Workshop on Language Technologies and Computational Social Science}, \bibinfo{publisher}{ACL}. pp. \bibinfo{pages}{18--22}.
\bibitem[{Williams et~al.(2020)Williams, Rodrigues and Novak}]{williams2020accenture}
\bibinfo{author}{Williams, E.}, \bibinfo{author}{Rodrigues, P.}, \bibinfo{author}{Novak, V.}, \bibinfo{year}{2020}.
\newblock \bibinfo{title}{Accenture at checkthat! 2020: If you say so: Post-hoc fact-checking of claims using transformer-based models} \href{http://arxiv.org/abs/2009.02431}{{\tt arXiv:2009.02431}}.
\bibitem[{Williams et~al.(2021)Williams, Rodrigues and Tran}]{williams2021accenture}
\bibinfo{author}{Williams, E.}, \bibinfo{author}{Rodrigues, P.}, \bibinfo{author}{Tran, S.}, \bibinfo{year}{2021}.
\newblock \bibinfo{title}{Accenture at checkthat! 2021: interesting claim identification and ranking with contextually sensitive lexical training data augmentation}.
\newblock \bibinfo{journal}{arXiv:2107.05684} .
\bibitem[{Yang et~al.(2019)Yang, Dai, Yang, Carbonell, Salakhutdinov and Le}]{yang2019xlnet}
\bibinfo{author}{Yang, Z.}, \bibinfo{author}{Dai, Z.}, \bibinfo{author}{Yang, Y.}, \bibinfo{author}{Carbonell, J.}, \bibinfo{author}{Salakhutdinov, R.R.}, \bibinfo{author}{Le, Q.V.}, \bibinfo{year}{2019}.
\newblock \bibinfo{title}{Xlnet: Generalized autoregressive pretraining for language understanding}, in: \bibinfo{booktitle}{Proc. of NeurIPS}, pp. \bibinfo{pages}{5753--5763}.
\bibitem[{Zhang and Ghorbani(2020)}]{zhang2020overview}
\bibinfo{author}{Zhang, X.}, \bibinfo{author}{Ghorbani, A.A.}, \bibinfo{year}{2020}.
\newblock \bibinfo{title}{An overview of online fake news: Characterization, detection, and discussion}.
\newblock \bibinfo{journal}{Information Processing \& Management} \bibinfo{volume}{57}, \bibinfo{pages}{102025}.
\bibitem[{Zhou et~al.(2021)Zhou, Wu and Fung}]{zhou2021fight}
\bibinfo{author}{Zhou, X.}, \bibinfo{author}{Wu, B.}, \bibinfo{author}{Fung, P.}, \bibinfo{year}{2021}.
\newblock \bibinfo{title}{Fight for 4230 at checkthat! 2021: Domain-specific preprocessing and pretrained model for ranking claims by check-worthiness.}, in: \bibinfo{booktitle}{CLEF (Working Notes)}, pp. \bibinfo{pages}{681--692}.

\end{thebibliography}
 
\end{document}